\documentclass[sigconf]{acmart}

\usepackage{booktabs} 
\usepackage{url}  
\usepackage{graphicx}  
\usepackage{amsmath}
\usepackage{subcaption}
\usepackage{multirow}
\usepackage{array}
\newcolumntype{P}[1]{>{\centering\arraybackslash}p{#1}}
\newcolumntype{M}[1]{>{\centering\arraybackslash}m{#1}}
\setcopyright{rightsretained}

\begin{document}
\title{Describing Natural Images Containing Novel Objects with \\
       Knowledge Guided Assitance}

\author{Aditya Mogadala}
\affiliation{%
  \institution{Karlsruhe Institute of Technology}
  \city{Karlsruhe} 
  \state{Germany} 
}
\email{aditya.mogadala@kit.edu}

\author{Umanga Bista}
\affiliation{%
  \institution{Australian National University}
  \city{Canberra} 
  \state{Australia} 
}
\email{umanga.bista@anu.edu}

\author{Lexing Xie}
\affiliation{%
  \institution{Australian National University}
  \city{Canberra} 
  \country{Australia}
  }
\email{lexing.xie@anu.edu}

\author{Achim Rettinger}
\affiliation{%
  \institution{Karlsruhe Institute of Technology}
  \city{Karlsruhe}
  \country{Germany}
}
\email{rettinger@kit.edu}

\renewcommand{\shortauthors}{A. Mogadala et al.}

\begin{abstract}
Images in the wild encapsulate rich knowledge about varied abstract concepts and cannot be sufficiently described with models built only using image-caption pairs containing 
selected objects. We propose to handle such a task with the guidance of a knowledge base that incorporate many abstract concepts. Our method is a two-step process where we 
first build a multi-entity-label image recognition model to predict abstract concepts as image labels and then leverage them in the second 
step as an external semantic attention and constrained inference in the caption generation model for describing images that depict unseen/novel objects. Evaluations show 
that our models outperform most of the prior work for out-of-domain captioning on MSCOCO and are useful for integration of knowledge and vision in general.
\end{abstract}

%



\maketitle

\section{Introduction}
\label{sec:intro}
Content on the Web is highly heterogeneous comprising mostly visual and textual modalities. In most cases, these modalities complement each other to illustrate the 
semantics of concepts and objects. Many approaches have leveraged such multi-view content for grounding textual with visual information or vice versa. Natural 
language processing (NLP) tasks like monolingual word similarity~\cite{lazaridou2015} and language learning~\cite{chrupala2015} were improved by grounding textual with visual information. The 
grounding of visual with textual information has shown improvement for existing computer vision (CV) tasks such as image annotation~\cite{srivastava2012}. The aforementioned tasks can be further 
subdivided into two broad categories where the first category of tasks combines the semantics of multiple modalities to achieve common representation for representational 
tasks such as generation of descriptions for images or videos~\cite{venugopalan2014,vinyals2017} and visual question answering~\cite{antol2015}. While the second category of 
tasks leverages cross-modal semantics to identify the relationship between visual and textual content for attaining referential grounding~\cite{boleda2017}.

Several approaches are proposed to solve varied tasks separately with methods from both categories. However, procedures from both categories can benefit by complementing 
each other. For example, many approaches for the representational task of generating captions for images are inspired by encoder-decoder 
architecture or its variations that encompass an attention mechanism~\cite{bahdanau2014} learn model using only (visual)image-(textual)caption parallel corpora. This causes the 
model to fail in describing those images which contain unseen/novel objects and concepts not part of parallel captions. Also, the vocabulary is limited to frequent words in the captions and 
often fails to incorporate rare or infrequent words. Given such challenges, we strive towards a solution that addresses the bottleneck in the aforementioned representational task. 

Usage of structured information provided by a knowledge base (KB)~\cite{lehmann2015} has shown to assist several textual tasks such as question answering over 
structured data~\cite{bordes2014}, language modeling~\cite{ahn2016}, and generation of factoid questions~\cite{serban2016}. Our hypothesis is that caption generation for 
images containing unseen/novel objects can significantly benefit from employing structured information (henceforth called knowledge) provided by a KB.

Thus in this paper, we aim to address the task of generating captions that include unseen visual objects in images with our proposed solution termed as knowledge guided assistance (KGA). The 
aim of KGA is to operate as referential grounding by providing external semantic attention during training and also work as a dynamic constraint while inference of a 
caption generation model. In particular, KGA assists to generate captions for unseen/novel objects in images which lack parallel captions. This makes KGA differ 
from earlier approaches that perform similar task such as deep compositional captioning (DCC)~\cite{anne2016} and novel object captioner (NOC)~\cite{venu2017} by not 
depending solely on the corpus specific word semantics for next word prediction in a caption generation model. Also, when compared with constrained beam 
search (CBS)~\cite{anderson2017} KGA incorporates more information from textual caption data by not constraining solely on image tags. In this respect, KGA is closer 
to LSTM-C~\cite{yao2017} which uses object classifiers and copying mechanism, but KGA diverges by using attention mechanism and dynamic weight transfer as opposed to word 
copying. Figure~\ref{fig:prob} presents an overview and the main contributions of this paper are summarized as follows:
\begin{figure*}[!ht]
  \centering{}
    \includegraphics[width=0.65\textwidth]{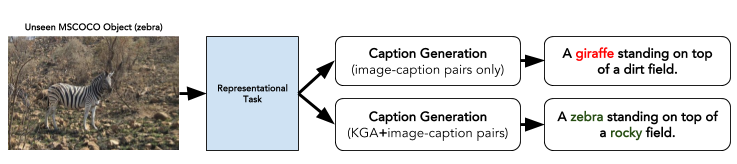}
  \caption{Intention of KGA for representational task of unseen/novel image object caption generation. }
  \label{fig:prob}
\end{figure*}
\begin{itemize}
 \item We designed a novel approach to improve the representational task of caption generation including unseen/novel visual objects with the assistance from a knowledge base. 
 \item We created a multi-label image classifier for grounding depicted visual objects to knowledge base entities. 
 \item We conducted an extensive experimental evaluation showing effectiveness of KGA.
\end{itemize}
\section{Related Work}
Our related work can be drawn from many closely aligned areas.
\subsection{Grounding Natural Language in Images}
The grounding of natural language in images is employed to comprehend objects and their relationships. Flickr30k Entities~\cite{plummer2015} is one such approach that augments 
the Flickr30k dataset images with bounding boxes using all noun phrases present in their parallel textual descriptions. We also leveraged textual descriptions for grounding in images, 
but to explicitly relate objects with their knowledge base entities. Other approaches~\cite{feres2015,johnson2015} also tried to relate knowledge to images, but not by explicitly linking it to 
KB. However to extract visual knowledge for supporting tasks such as question answering and image retrieval.
\subsection{Attention Mechanism in Caption Generation}
Initially, attention mechanism was applied to tasks such as image caption generation~\cite{xu2015} with two different possibilities i.e. soft and hard attention. 
Another recent improvement is seen in performing adaptive attention with visual sentinel~\cite{lu2016} by identifying when to look inside an 
image for cues and not every time as aforementioned would do it. As both of the aforementioned approaches would look entire image to add global context, region-based attention~\cite{jin2015} was 
proposed to experience visual perception where the attention shifting among the visual regions imposes a thread of visual ordering. Slightly, deviating from earlier visual feature centric 
approaches, the attribute based attention~\cite{you2016} extracted attributes from an image and used them as input vectors resembles our approach.
\subsection{Dealing with Rare/OoV Words}
Usage of external vocabulary or the structured data is becoming prominent in many neural network models. Goal of these approaches is 
to copy information from external sources whenever the neural network models fail to predict with certain confidence. Recently, a neural knowledge language model~\cite{ahn2016} was 
proposed to improve language modeling with external structured data to improve tasks which are dependent on entities and also extended for text generation~\cite{lebret2016}. 
Some approaches~\cite{gu2016} have used copying mechanism to deal with out-of-vocabulary (OoV) words, while few adopted pointing mechanism~\cite{merity2016}. Our approach for constrained 
inference fall in-line with such approaches, but rather prefer to enhance neural model weights than direct copying.
\section{Describing Images with Novel Objects Using Knowledge Guided Assistance (KGA)}
In this section, we present our caption generation model for generating captions for unseen/novel image objects with support from KGA. The core goal of KGA is to introduce 
external semantic attention (ESA) into the learning caption generation models and KGA also work as a constraint for transferring learned weights between seen and unseen 
semantic and word image labels during inference.
\subsection{Caption Generation Model}
\label{sec:cap}
\begin{figure*}[!ht]
  \centering{}
    \includegraphics[width=0.8\textwidth]{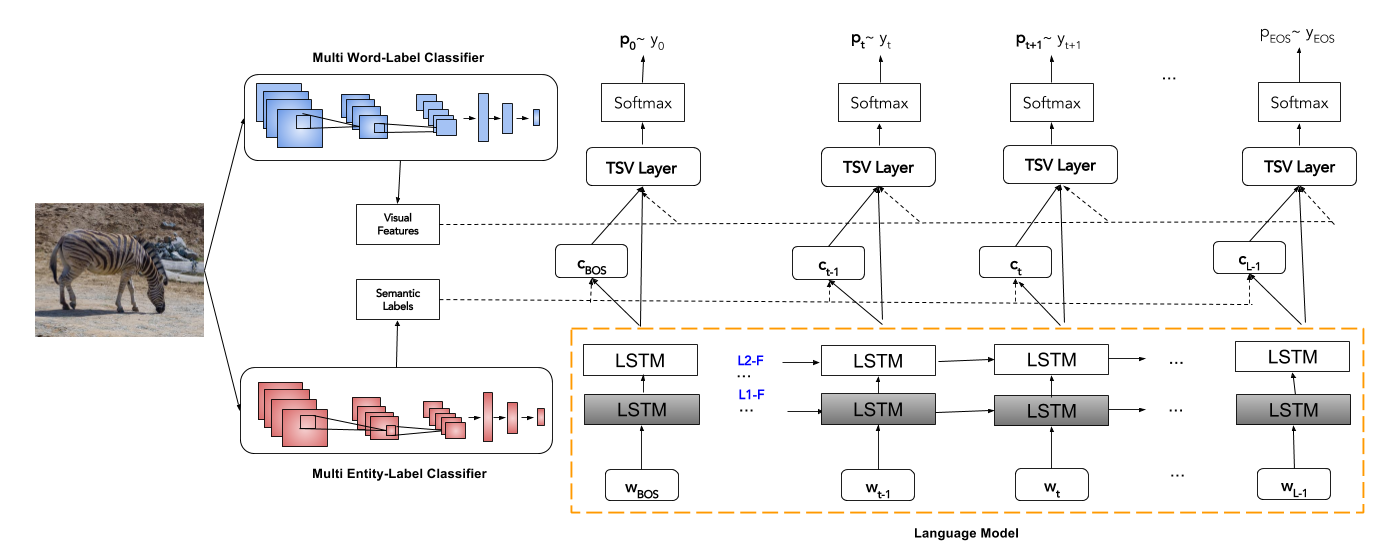}
  \caption{Our caption generation model (KGA-CGM) built with three components. A language model implemented with a 2-layer forward LSTM where L1-F and L2-F represents 
  layer-1 and layer-2 respectively, a multi-word-label classifier to generate image visual features and multi-entity-label classifier to support ESA. $w_t$ represents the input caption word, $c_t$ the semantic attention, 
  $p_{t}$ the output of probability distribution over all words and $y_t$ the predicted word at each time step $t$. BOS and EOS represent the special beginning and end of sentence tokens respectively.}
  \label{fig:model}
\end{figure*}

Our image caption generation model (henceforth, KGA-CGM) combines three important components: a language model pre-trained on unpaired 
textual corpora, external semantic attention (ESA) and image features with a textual (T), semantic (S) and visual (V) layer (i.e. TSV layer) for predicting the next word in 
the sequence when learned using image-caption pairs. In the following, we present each of these components separately while Figure~\ref{fig:model} presents the overall 
architecture of KGA-CGM.
\subsubsection{Language Model}
This component is crucial to transfer the sentence structure for unseen visual objects. Language model is implemented with two long short-term 
memory (LSTM)~\cite{hochreiter1997} layers to predict the next word given previous words in a sentence. If $\overrightarrow{\boldsymbol{w}_{1:L}}$ represent 
the input to the forward LSTM of layer-1 for capturing forward input sequences into hidden sequence vectors ($\overrightarrow{\boldsymbol{h}_{1:L}^1} \in \mathcal{R}^{H}$), 
where $L$ is the final time step. Then encoding of input word sequences into hidden layer-1 sequences at each time step $t$ is achieved as follows:
\begin{equation}
 \overrightarrow{\boldsymbol{h}_{t}^1} = \text{L1-F}(\overrightarrow{\boldsymbol{w}_t};\Theta)
\end{equation}
Similarly, layer-1 hidden sequences supplied as input to layer-2 are encoded as follows:
\begin{equation}
 \overrightarrow{\boldsymbol{h}_{t}^2} = \text{L2-F}(\overrightarrow{\boldsymbol{h}_{t}^1};\Theta)
\end{equation}
where $\Theta$ represent hidden layer parameters. Finally, the encoded hidden sequence ($\overrightarrow{\boldsymbol{h}_{t}^2} \in \mathcal{R}^{H}$) at time step $t$ is then used for predicting the probability 
distribution of the next word given by Equation~\ref{eqn:pt}. 
\begin{equation}
\label{eqn:pt}
 \boldsymbol{p}_{t+1} = softmax(\boldsymbol{h}_{t}^2)
\end{equation}
The softmax layer is only used while training with unpaired textual corpora and not used when learned with image captions.
\subsubsection{External Semantic Attention (ESA)}
Attention mechanism was first introduced by Bahdanau et al.~\shortcite{bahdanau2014} in an encoder-decoder architecture. It was particularly useful for dynamically 
changing context while decoding. Later, the attention mechanism was extended to other tasks such as speech recognition~\cite{chorowski2015} and multimodal representational 
tasks. A feasible case of multimodal representational task is image caption generation, where attention is perceived from two different aspects i.e. visual attention~\cite{xu2015} and attribute 
attention~\cite{you2016}. Visual attention leveraged spatial image features, while attribute attention leveraged attributes obtained from an image.

Our objective from ESA is similar to attribute attention, but leverages entity annotations as semantic labels obtained using a multi-entity-label image 
classifier (discussed in the later section). Entity annotations obtained are analogous to image patches and attributes.

In formal terms, if $\boldsymbol{ea}_i$ is an entity annotation label and $\boldsymbol{e}_i \in \mathcal{R}^E$ the entity annotation vector among set of entity annotation 
vectors ($i=1,..,L$) and $\boldsymbol{\beta}_{i}$ the attention weight of $\boldsymbol{e}_i$ then $\beta_i$ is calculated at each time step $t$ using 
Equation~\ref{eqn:attsw}.
\begin{equation}
\label{eqn:attsw}
\boldsymbol{\beta}_{ti} = \frac{exp(\boldsymbol{p}_{ti})}{\sum_{j=1}^L exp(\boldsymbol{p}_{tj})}
\end{equation}
\begin{equation}
\label{eqn:pt1}
\boldsymbol{p}_{ti} = f(\boldsymbol{e}_i,\boldsymbol{h}_t)
\end{equation}
where $\boldsymbol{p}_{ti}$ given by Equation~\ref{eqn:pt1} and $f(\boldsymbol{e}_i,\boldsymbol{h}_t)$ represent scoring function which conditions on hidden state 
($\boldsymbol{h}_{t}$) of a caption language model. It can be observed that the scoring function 
$f(\boldsymbol{e}_i,\boldsymbol{h}_t)$ is crucial for deciding attention weights. Also, relevance of the hidden state with each entity annotation is calculated using 
Equation~\ref{eqn:eprevw}.
\begin{equation}
\label{eqn:eprevw}
f(\boldsymbol{e}_i,\boldsymbol{h}_t) = tanh(\boldsymbol{h}_{t}^TW_{he}\boldsymbol{e}_i)
\end{equation}
where $\boldsymbol{h}_t \in \mathcal{R}^H$, $W_{he} \in \mathcal{R}^{H \times E}$ is a bilinear parameter matrix. Once the attention weights are calculated, the soft attention weighted annotation vector of the context $\boldsymbol{c}$, which is a dynamic representation of the caption at time step $t$ is 
given by Equation~\ref{eqn:sasa}
\begin{equation}
\label{eqn:sasa}
\boldsymbol{c}_t = \sum^{L}_{i=1} \beta_{ti}\boldsymbol{e}_i
\end{equation}
Here, $\boldsymbol{c}_t \in \mathcal{R}^E$ and $L$ represent the cardinality of entity class annotations per image-caption pair instance.
\subsubsection{Image Features}
The image aligned with captions are used to extract visual features using multi-word-label image classifier (discussed more in later sections). To be consistent with 
other approaches~\cite{anne2016,venu2017} and for a fair comparison, our visual features ($I$) also have each index corresponding to the probability of word-label annotation in the image.
\subsubsection{TSV layer}
Once the output from all components is acquired, the TSV layer is employed to integrate their features i.e. textual ($T$), semantic ($S$) and visual ($V$) yielded by language model, ESA and 
images respectively. Thus, TSV acts as a transformation layer for molding three different feature spaces into a single common space for prediction of next word in the sequence. 

If $\boldsymbol{h}_{t}^2 \in \mathcal{R}^H$, $\boldsymbol{c}_t \in \mathcal{R}^E $  and $\boldsymbol{I}_t \in \mathcal{R}^I$ represent vectors acquired at each time step $t$ from 
language model, ESA and images respectively. Then the integration at TSV layer of KGA-CGM is provided by Equation~\ref{eqn:mer}.
\begin{equation}
\label{eqn:mer}
 \boldsymbol{TSV}_{t} = W_{\boldsymbol{h}_{t}^2}\boldsymbol{h}_{t}^2+W_{\boldsymbol{c}_t}\boldsymbol{c}_t+W_{\boldsymbol{I}_t}\boldsymbol{I}_t
\end{equation}
where $W_{\boldsymbol{h}_{t}^2} \in \mathcal{R}^{vs \times H}, W_{\boldsymbol{c}_t} \in \mathcal{R}^{vs \times E}$ and $W_{\boldsymbol{I}_t} \in \mathcal{R}^{vs \times I}$ are 
linear conversion matrices and $\boldsymbol{vs}$ is the image-caption pair training dataset vocabulary size.
\subsubsection{Word Prediction}
The output from the TSV layer at each time step $t$ is further used for predicting the next word in the sequence using a softmax layer given by Equation~\ref{eqn:soft}.
\begin{equation}
\label{eqn:soft}
 \boldsymbol{p}_{t+1} = softmax(\boldsymbol{TSV}_t)
\end{equation}

\subsection{KGA-CGM Training}
To learn parameters of KGA-CGM, first we freeze the parameters of the language model trained using unpaired textual corpora. Thus, enabling only those parameters to be learned with 
image-caption pairs emerging from ESA and TSV layer such as $W_{he},W_{\boldsymbol{h}_{t}^2},W_{\boldsymbol{c}_t}$ and $W_{\boldsymbol{I}_t}$.  KGA-CGM is now trained to optimize the cost function that minimizes the sum of the negative log likelihood of the appropriate word at each time step given by Equation~\ref{eqn:logloss}.
\begin{equation}
\label{eqn:logloss}
 \min_{\theta} -\frac{1}{N}\sum_{n=1}^N \sum_{t=0}^{L^{(n)}} log (\boldsymbol{p}(y^{(n)}_t))
\end{equation}
Where $L^{(n)}$ represent the length of sentence (i.e. caption) with beginning of sentence (BOS), end of sentence (EOS) tokens at $n$-th training sample and $N$ as a number of samples used for 
training. 
\subsection{KGA-CGM Constrained Inference}
Inference is not straightforward as in earlier image caption generation approaches~\cite{vinyals2017}. Unseen/novel image objects have no parallel captions throughout 
training, hence they will never be generated in a caption during inference. Thus, unseen/novel objects require guidance from similar objects or external sources during inference. Earlier approaches such as DCC~\cite{anne2016} have leveraged similar concepts (i.e image word-labels) to transfer weights between seen and unseen word-labels. However, similar labels are 
found only using word embeddings of textual corpora and are not constrained on images. This obstructs the view from an image leading to spurious results. We resolve such issues by constraining 
the weight transfer between seen and unseen image labels (i.e. both semantic and word) with help from KGA. As the first step, we identify closest similar word-label of unseen objects with their Glove embeddings~\cite{pennington2014} learned using unpaired textual corpora. Now, for transferring 
weights between seen and unseen image labels, we perform dynamic weight transfer during test image caption generation with help of entity annotation semantic labels ($\boldsymbol{ea}$) provided by 
multi-entity-label image classifier. Whenever the word predicted by our KGA-CGM model is the closest similar word of an unseen object then the unseen object is checked for its presence in 
$\boldsymbol{ea}$. If the unseen object is part of $\boldsymbol{ea}$, then direct transfer of weights is performed between seen and unseen with 
$W_{\boldsymbol{c}_t}$[unseen,:]=$W_{\boldsymbol{c}_t}$[closest,:] and $W_{\boldsymbol{I}_t}$[unseen,:]=$W_{\boldsymbol{I}_t}$[closest,:] and then setting $W_{\boldsymbol{I}_t}$[unseen,closest],
$W_{\boldsymbol{I}_t}$[closest,unseen]=0 to remove mutual dependencies 
of their presence in an image. For the next image test sample, weights are again set back to their initial states. Beam search is used to consider the best 
$k$ sentences at time $t$ to identify the sentence at next time step. In this research, we perform experiments with $k$=1 and $k$=3.
\section{Learning Multi-Label Image Classifiers}
\label{sec:multicl}
It can be perceived from earlier sections that the important constituents that influence KGA are the image semantic labels and features. Image features embody 
objects/actions/scenes identified in an image, while semantic labels capture the external semantic attention and entity class labels. In this section, we present the approach to extract both 
image features and semantic labels.
\subsection{Multi-Word-label Image Classifier}
\label{ssec:mwlc}
To extract image features, emulating Hendricks et al.~\cite{anne2016} a multi-word-label classifier is built using the caption aligned to an image by extracting 
part-of-speech (POS) tags such as nouns, verbs and adjectives attained for each word. For example, the caption ``A young child brushes his teeth at the sink'' contains word-labels 
such as ``young (JJ)'', ``child (NN)'', ``teeth (NN)'' etc., that represent abstract concepts in an image. An image classifier is trained now with multiple word-labels using a sigmoid 
cross-entropy loss by fine-tuning VGG-16~\cite{simonyan2014} pre-trained on the training part of the ILSVRC-2012.
\subsection{Multi-Entity-label Image Classifier}
\label{ssec:melc}
To extract semantic labels which are analogous to the word-labels, the multi-entity-label classifier is build with entity labels attained from a knowledge base annotation tool such as 
DBpedia spotlight\footnote{\url{https://github.com/dbpedia-spotlight/}}. Considering the caption presented in the aforementioned section, entities extracted from the 
caption are ``Brush\footnote{\url{http://dbpedia.org/resource/Brush}}'' and ``Tooth\footnote{\url{http://dbpedia.org/resource/Tooth}}'', which will be treated as semantic labels. An image 
classifier is now trained with multiple entity-labels using sigmoid cross-entropy loss by fine-tuning VGG-16~\cite{simonyan2014} pre-trained on the training part of the ILSVRC-2012. 
\section{Experimental Setup}
In this section, we describe the experimental setup employed for our experiments.
\subsection{Resources and Datasets}
Our approaches are dependent on several resources such as toolkits etc., while datasets are utilized to conduct evaluations. In the following, resources, datasets and evaluation 
measures are presented.
\subsubsection{Knowledge Bases (KBs)}
There are several openly available KBs such as DBpedia\footnote{\url{http://wiki.dbpedia.org/}}, Wikidata\footnote{\url{https://www.wikidata.org/wiki/Wikidata:Main_Page}}, and 
YAGO\footnote{\url{http://www.mpi-inf.mpg.de/departments/databases-and-information-systems/research/yago-naga/yago/downloads/}} which cover general knowledge about entities and their 
relationships. We choose DBpedia as our KB for entity annotation, as it is one of the extensively used resource for semantic annotation and disambiguation~\cite{lehmann2015}.
\subsubsection{Unseen Objects in Image-Caption Dataset}
\label{ssec:hoicd}
To evaluate KGA-CGM, we use the subset of the MSCOCO~\cite{lin2014} dataset proposed by Hendricks et al.~\shortcite{anne2016}. The dataset is obtained by clustering 80 image object 
category labels into 8 clusters and then selecting one object from each cluster to be held out from the training set.  Now the training set does not contain the images and sentences 
of those 8 objects represented by bottle, bus, couch, microwave, pizza, racket, suitcase and zebra. Thus making the MSCOCO training dataset to constitute 70,194 image-caption pairs. While 
validation set of 40504 image-caption pairs are again divided into 20252 each for testing and validation. Now, the goal of KGA-CGM is to generate caption for those test images which contain 
these 8 unseen object categories.
\subsection{Evaluation Measures}
To evaluate the representational task of caption generation, we use evaluation metrics same as earlier approaches~\cite{anne2016,venu2017,yao2017} such as METEOR and also 
SPICE~\cite{anderson2017} to check the effectiveness of generated caption. CIDEr~\cite{vedantam2015} metric is not used as it is required to calculate inverse document frequency used by this metric across the 
entire test set and not just unseen object subsets. F1-score is also calculated to measure the presence of unseen objects in the generated captions when compared against reference captions.
\section{Experiments}
The experiments are conducted to evaluate the efficacy of the proposed approaches and their dependencies.
\subsection{Implementing Multi-Label Classifiers}
The key component of unseen image object caption generation is comprehending visual information. To realize it, we re-use existing and also build our own image multi-label classifier. 
\subsubsection{Multi-Word-Label Classifier}
The principal role of the multi-word-label classifier is to provide image features for caption generation. We use the pre-trained model~\cite{anne2016} 
trained on the subset of MSCOCO using the approach presented earlier. The image features extracted represent the probabilities of 471 image labels occurring in a given image.
\subsubsection{Multi-Entity-label Classifier}
The goal of multi-entity-label classifier is to recognize multiple semantic labels per image. To build the classifier, training set of MSCOCO dataset constituting 82,783 training 
image-caption pairs are used to extract around 812 unique labels with an average of 3.2 labels annotated per image. Additionally, features for the images are extracted using different layers such as pool5, fc6 and fc7 of VGG-16~\cite{simonyan2014} pre-trained on ILSVRC-2012 for fine-tuning with our 
semantic labels. Also, to comprehend the contribution of aforementioned layers to the accuracy of the classifier, we throughly analyzed separately the classifiers built using pool5, fc6 and 
fc7 as the initialization layers before fine-tuning. Our analysis revealed that pool5 features overfit even with regularization. 

\begin{figure}[!ht]
    \centering
    \includegraphics[width=0.35\textwidth]{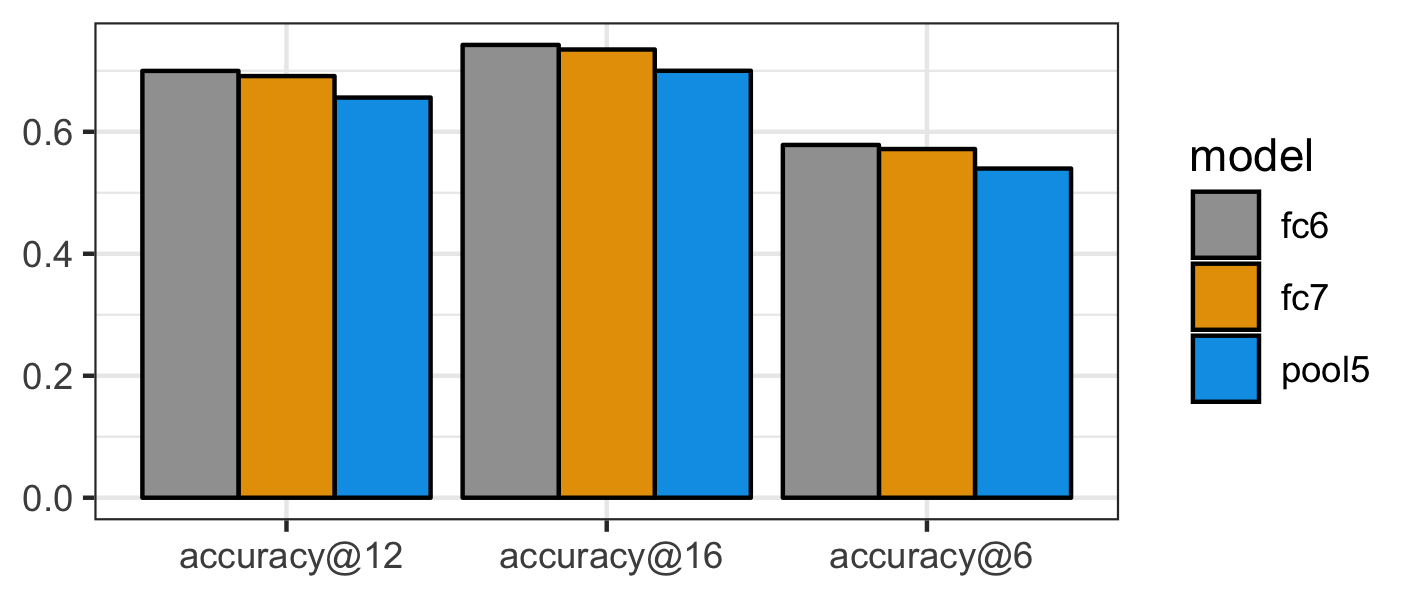}
    \caption{Accuracy of the predicted labels on the test set by Multi Entity-Label Classifier.}
    \label{fig:test_recall}
\end{figure}

To address this challenge, we trained a classifier with Caffe\footnote{\url{http://caffe.berkeleyvision.org/}} by fine-tuning the layers above fc6 and fc7 which gave us an 
improvement in the accuracy as observed in Table~\ref{table:params}.
\begin{table}
    \centering
    \caption{Validation results of different VGG-16 layers. Hyper parameters are used to fine-tune Caffe VGG-16 model. Accuracy@K is calculated by predicting as many labels as in ground truth 
     for each image.}
     \label{table:params}
    \begin{tabular}{llll}
         \hline
         \multicolumn{4}{c}{Model} \\
         \hline
         Hyper Parameters & pool5 & fc6 & fc7\\
         \hline
         weight\_decay &    0.05 &    0.03 &    0.01\\
         base\_lr      &   0.001 &  0.0003 &   0.003\\
         gamma         &     0.5 &     0.5 &    0.33\\
         stepsize      &    7.5K &     10K &      8K\\
         maxiter       &     60K &     50K &     40K\\
         momentum      &     0.9 &     0.9 &    0.9 \\
         batch\_size   &     256 &     256 &    256 \\
         \hline
         \multicolumn{4}{c}{Results} \\
         \hline
         Validation Loss        & 11.0035 & \textbf{10.1152} & 10.3372\\
         Accuracy@12 &  0.6572 & \textbf{0.7018}  & 0.6868\\
         Accuracy@K  &  0.4526 & \textbf{0.4892}  &  0.4778\\
         \hline
    \end{tabular}
\end{table}
The classifier fine-tuned on fc6 features constitute two fully connected layers of 4096 and 812 neurons above fc6, with the first layer having 50\% dropout and ReLU activation while the 
output layer comprises a sigmoid activation. Similarly, the classifier fine-tuned with fc7 features have an output layer of 812 neurons comprising a sigmoid activation. The loss 
function used during training is sigmoid cross-entropy, while only sigmoid is used during prediction for exhibiting the presence of label probabilities.
Figure~\ref{fig:test_recall} shows the predictions on the test dataset. It can be observed that fc6 gave the best result with an accuracy around 70\% for top-12 and 74.4\% top-16. 
\subsection{Entity-Label Embeddings}
\label{ssec:ele}
We presented that the acquisition of labels for multi-entity-label classifiers was obtained by the DBpedia spotlight entity annotation and disambiguation 
tool. These labels (i.e. entities) are expected to encapsulate general knowledge (e.g. encyclopedic knowledge) which is inter linked. Approaches~\cite{socher2013a} earlier have transformed such entities in a knowledge base into embeddings to capture their relational information for tasks such as knowledge base 
completion. In our research as well, we see the efficacy of these embeddings for caption generation. Entity embeddings leverages external semantic information to be used for attention. To 
obtain entity-label embeddings, we adopted the RDF2Vec~\cite{ristoski2016} approach and generated 500 dimensional vector representations for all 812 labels used to represent images in the 
entire MSCOCO. 

We further qualitatively evaluate these entity-label embeddings to check there affect on caption generation. Most images are respresented with more than one entity-label, thus 
providing multi-label information for each image. However, directly using their embeddings for ESA can affect caption generation if the label embeddings are not closely related. To check 
for their closely relatedness, we perform entity similarity. Table~\ref{unseenclose} shows the results of uneen/novel mscoco objects.
\begin{table}[!ht]
 \centering
 \small
 \caption{Top-5 closely related entities of  unseen MSCOCO Objects}
 \label{unseenclose}
 \begin{tabular}{ l|l}
 \hline
 Unseen Object & Top-5 Closely Related Entities\\
 \hline
 Bottle & Wine\_bottle, Wine\_glass, Table\_setting \\
	&  Nap\_(textile), Tablecloth \\
\cline{2-2}
 Bus &  Truck, Double-decker\_bus, Transit\_bus \\
	 & Cargo, Tram \\
\cline{2-2}
 Couch & Pillow, Cupboard, Bathtub \\ 
       & Hair\_dryer, Living\_room\\
\cline{2-2}
 Microwave & Blender, Oven, Paper\_bag \\
	 &  Dishwasher, Refrigerator\\
\cline{2-2}
 Pizza & Pasta, Pepperoni, Salad \\ 
       &  Sauce, Grilling \\
\cline{2-2}
 Racket & Ball, Flying\_disc, Snowboard \\
        & Glove, Cricket\_ball\\
\cline{2-2}
 Suitcase & Baggage, Backpack, Hair\_dryer \\
          & Apron, Bathtub\\
\cline{2-2}
 Zebra & Giraffe, Elephant, Horn\_(anatomy) \\
       & Calf, Ox\\
 \hline
 \end{tabular}
\end{table}

It can be perceived from the Table~\ref{unseenclose} that most of the closely related entities always co-occur in an image as shown with few examples in the paper. Thus enhancing the caption 
generation model with ESA proven to be effective. 

\begin{figure*}[!ht]
    \centering
    \includegraphics[width=\textwidth]{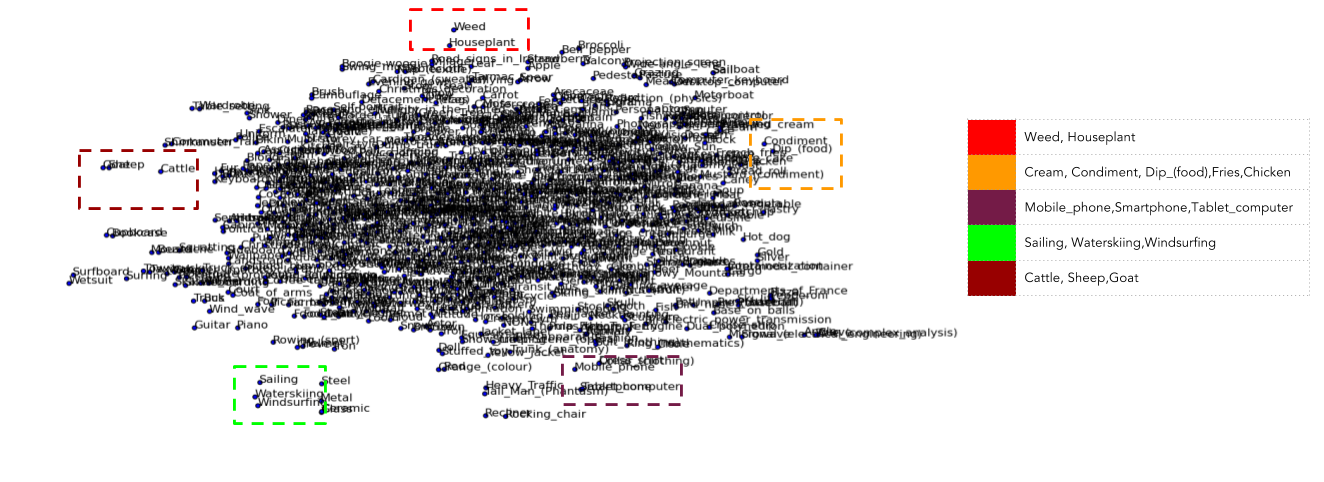}
    \caption{t-sne visualization of the entity-label embeddings.}
    \label{fig:tsne}
\end{figure*}

We also performed t-SNE visualization of all entity-labels to check how cluster together. It can be seen from the Figure~\ref{fig:tsne} visualization some of the closely related objects which occur in the 
same context cluster close to each other.

\subsection{Novel Objects Description}
In this section, we evaluate KGA-CGM for unseen/novel image objects caption generation.
\subsubsection{Implementation}
One of the important component of KGA-CGM model is language model. Even though its weights are fixed during learning, the words in a caption are initially set with pre-trained 
Glove~\cite{pennington2014} embeddings of 256 dimensions while the hidden layer dimensions are set to 512. Information about other components i.e. image features and semantic labels are 
already discussed in earlier sections. KGM-CGM is then trained with Adam optimizer with gradient clipping having maximum norm of 1.0 for about 15$\sim$50 epochs. Validation data 
is used for fine tuning parameters and model selection.
\subsubsection{Ablation Study}
To understand how different aspects of training KGA-CGM influence the unseen image objects caption generation, we perform ablation study by removing different components of 
KGA-CGM. Table~\ref{ablation} present the results obtained.  
\begin{table}[!ht]
 \centering
 \small
 \caption{Affect on KGA-CGM (Avg)}
 \label{ablation}
 \begin{tabular}{ l|l|l|l|l}
 \hline
 Model & Beam & METEOR & SPICE & F1\\
 \hline
 None  & 1 & 19.7 & 11.7  & 0  \\
 Only ESA & 1 & 20.5 & 12.8 & 0 \\
 Only CI & 1 & 20.1 & 12.3 & 39.8 \\
 ESA+CI & 1 & 22.2 & 14.6  & 54.5 \\
 ESA+CI & $>$1 & 21.5 & 13.9  & 48.9 \\
 \hline
 \end{tabular}
\end{table}
It can be noticed that \textit{None}, which refers to the no usage of either ESA or constrained inference (CI) in the KGA-CGM model have F1 measure 0. Enabling ESA into this basic caption 
generation model has shown an increase in the METEOR and SPICE as observed in \textit{Only ESA}. However, F1 measure has remained 0 due to no transfer of 
weights between seen and unseen image objects. Alternatively, enabling CI showed a jump in F1 measure as seen in \textit{Only CI}. However, both METEOR and SPICE are lower than \textit{Only ESA} 
due to missing weights for ESA. Enabling both ESA and CI make our complete KGA-CGM model equipped with both semantic attention from the image as well as constrained transfer of weights providing 
highest METEOR and SPICE scores of 22.2 and 14.6 respectively as observed in \textit{ESA+CI}. Also, it has increased F1 measure when compared to \textit{Only CI}. This shows that the 
coherent and accurately generated caption is important for presence of an object in caption. We also analyzed the effect of beam size on KGA-CGM and it can be observed that increasing beam 
size during inference has shown a drop in all measures. This can be attributed to the usage of terms which are outside unseen objects captions vocabulary and are more general to entire caption 
dataset.
\subsubsection{Quantitative Analysis}
We compared our complete KGA-CGM model with the other existing models that generated captions for the unseen MSCOCO image objects. To have a fair comparison, only those results are 
compared that used VGG-16 to generate image features. Table~\ref{dataset-f1-capgen} shows the comparison of average scores based on METEOR, SPICE and F1 on all 8 unseen 
image objects with beam size 1 and greater than that ($>1$). It can be noticed that KGA-CGM with beam size 1 was comparable to other approaches even though it used fixed vocabulary and image 
tags. For example, CBS~\cite{anderson2017} used expanded vocabulary of 21,689 when compared to 8802 by us. Also, our word-labels per image are fixed, while CBS uses a varying size of 
predicted image tags (T1-4). This makes it nondeterministic and can increase uncertainty, as varying tags will either increase or decrease the performance. In Table~\ref{indobj}, we also 
present individual scores from all 8 unseen objects separately. Though our average scores were comparable to other approaches, for some of the unseen objects we attain state of the art results. 
We also add more analysis about the important components used in the KGA-CGM model in the Appendix.
\begin{table*}[!ht]
 \centering
 \small
 \caption{Average measure of all 8 unseen objects of MSCOCO with beam size 1 and greater than that ($>1$)}
 \label{dataset-f1-capgen}
 \begin{tabular}{ l|l|l|l|l}
 \hline
 Model & Beam & METEOR & SPICE & F1-score \\
 \hline
 DCC~\cite{anne2016} & 1 & 21.0 & 13.4 & 39.7 \\
 NOC~\cite{venu2017}&  1 & 20.7 & - & 50.5 \\
 \textbf{KGA-CGM} & 1 & 22.2 & 14.6 & \underline{54.5} \\
 \cline{1-2}
 NOC~\cite{venu2017}& $>$1 & 21.3 & - & 48.8 \\
 LSTM-C~\cite{yao2017} & $>$1 & 23.0 & - & \textbf{55.6} \\
 CBS(T4)~\cite{anderson2017} & $>$1  & 23.3 & 15.9 & 54.0 \\
 \textbf{KGA-CGM} & $>$1 & 21.5 & 13.9 & 48.9 \\
 \hline
 \end{tabular}
\end{table*}
\begin{table*}[!ht]
 \centering
 \small
 \caption{Individual measures for all 8 unseen objects. Best results are highlighted, while underline shows second best.}
 \label{indobj}
 \begin{tabular}{ l|l|l|l|l|l|l|l|l|l|l}
 \multicolumn{11}{ c }{} \\
 \hline
 Metric & Model & Beam & bottle & bus & couch & microwave & pizza & racket & suitcase & zebra  \\
 \hline
   & DCC~\cite{anne2016}      & 1 & 4.6 & 29.8 & \underline{45.9} & 28.1 & 64.6 & 52.2 & 13.2 & 79.9 \\
   & \textbf{KGA-CGM} & 1     & \underline{26.4} & 54.2 & 42.1 & \textbf{50.9} & \underline{70.8} & \textbf{75.3} & 25.6 & \underline{90.7} \\
   \cline{2-3}
F1 & NOC~\cite{venu2017}& $>$1 & 17.7 & \underline{68.7} & 25.5 & 24.7 & 69.3 & 55.3 & 39.8 & 89.0\\
   &CBS(T4)~\cite{anderson2017}& $>$1 & 16.3 & 67.8 & \textbf{48.2} & 29.7 & \textbf{77.2} & 57.1 & \textbf{49.9} & 85.7\\
   &LSTM-C~\cite{yao2017} & $>$1 & \textbf{29.6}  & \textbf{74.4} & 38.7 & 27.8 & 68.1 & \underline{70.2} & \underline{44.7} & \textbf{91.4} \\
   & \textbf{KGA-CGM} & $>$1 & 22.2  & 42.5  & 34.4 & \underline{48.1} & 69.6  & 63.1  & 22.6  & 88.5  \\

 \hline
       & DCC~\cite{anne2016} & 1 & 18.1 & \textbf{21.6} & \underline{23.1} & 22.1 & \textbf{22.2} & 20.3 & \underline{18.3} & 22.3 \\
       & \textbf{KGA-CGM}    & 1 & \textbf{21.5} & 20.3 & 23.0 & \underline{22.6} & 21.4 & \textbf{27.0} & \textbf{18.7} & \textbf{22.8} \\
      \cline{2-3}
METEOR & NOC~\cite{venu2017}& $>$1 & 21.2 & \underline{20.4} & 21.4 & 21.5 & \underline{21.8} & \underline{24.6} & 18.0 & 21.8\\
       & \textbf{KGA-CGM}   & $>$1 & \underline{21.3} & 19.2 & \textbf{23.5} & \textbf{23.2} & 21.7 & 22.5 & 18.0 & \underline{22.5} \\
       
 \hline
 SPICE & \textbf{KGA-CGM} & 1 & 13.1 & 12.6  & 14.9  & 13.3 & 13.2 & 19.8 & 10.6 & 19.6 \\
       & \textbf{KGA-CGM} & $>$1 & 12.6 & 11.6 & 14.6 & 13.6 & 13.1 & 16.7 & 10.3 & 18.6 \\
 \hline
 \end{tabular}
\end{table*}
\subsubsection{Qualitative Analysis}
In the Table~\ref{kgareprpred}, sample predictions of our best KGA-CGM model is presented. It can be observed that entity-labels has shown an influence for caption generation. Since, 
entities as image labels are already disambiguated, it attained high similarity in the prediction of a word thus adding useful semantics.
\begin{table*}[!ht]
\centering
\small
\caption{Predictions of KGA-CGM compared to NOC~\cite{venu2017} on MSCOCO with beam size 1.}
\label{kgareprpred}
\begin{tabular}{M{1.8cm}|M{2.5cm}|M {3.0cm} |M{3.0cm} |M{4.5cm}}
\hline
Image & Unseen Object & NOC & Our Predicted Caption & Predicted Entity-Labels (Top-3)\\
\hline
 \includegraphics[height=1.3cm,width=0.7in]{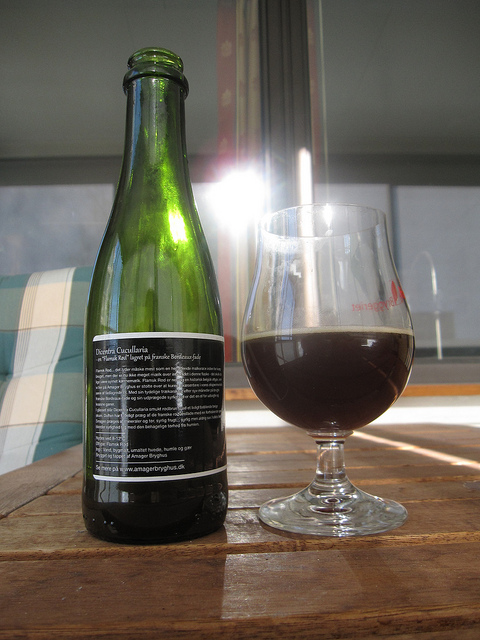}&  bottle & A wine bottle sitting on a table next to a wine bottle & A bottle of wine sitting on top of a table
 & Wine\_glass, Wine\_bottle, Bottle \\ 
\cline{2-5}
\includegraphics[height=1.25cm,width=0.7in]{bus} & bus & Bus driving down a street next to a bus stop. & A white bus is parked on the street & Bus,Public\_Transport,Transit\_Bus \\
\cline{2-5}
\includegraphics[height=1.25cm,width=0.7in]{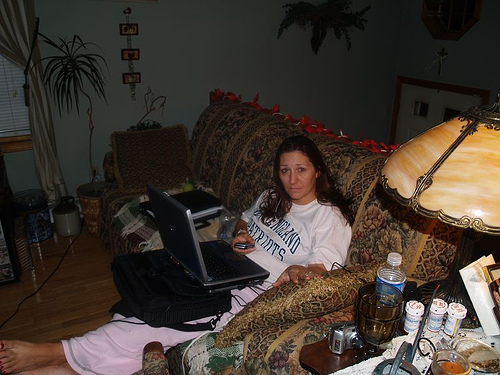} & couch & A woman sitting on a chair with a large piece of cake on her arm & A woman sitting on a couch with a remote & Cake,Couch,Glass \\ 
\cline{2-5}
\includegraphics[height=1.25cm,width=0.7in]{microwave} & microwave & A kitchen with a refrigerator, refrigerator, and refrigerator.& A kitchen with a microwave, oven and a refrigerator & Refrigerator,Oven,Microwave\_Oven \\
\cline{2-5}
\includegraphics[height=1.25cm,width=0.7in]{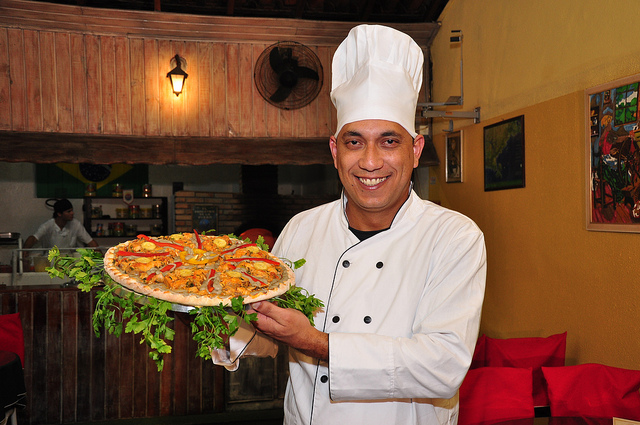} & pizza & A man standing next to a table with a pizza in front of it. & A man is holding a pizza in his hands & Pizza,Restaurant,Hat \\
\cline{2-5}
\includegraphics[height=1.25cm,width=0.7in]{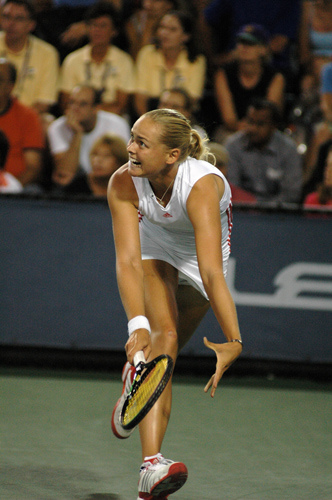} & racket & A woman court holding a tennis racket on a court & A woman playing tennis on a tennis court with a racket & Tennis, Racket\_(sports\_equipment), Court \\
\cline{2-5}
\includegraphics[height=1.25cm,width=0.7in]{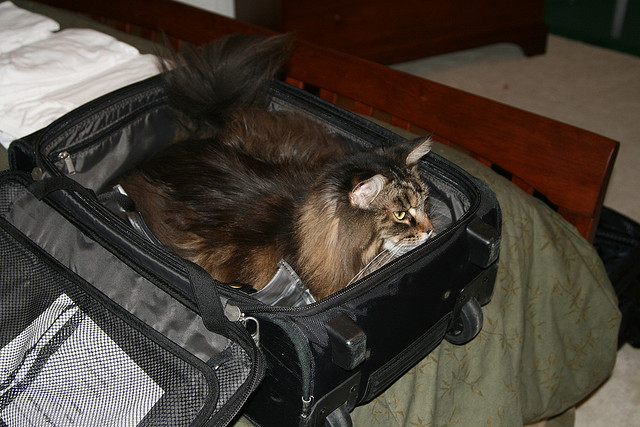} & suitcase & A cat laying on a suitcase on a bed. & A cat laying inside of a suitcase on a bed & Cat,Baggage,Black\_Cat \\
\cline{2-5}
\includegraphics[height=1.25cm,width=0.7in]{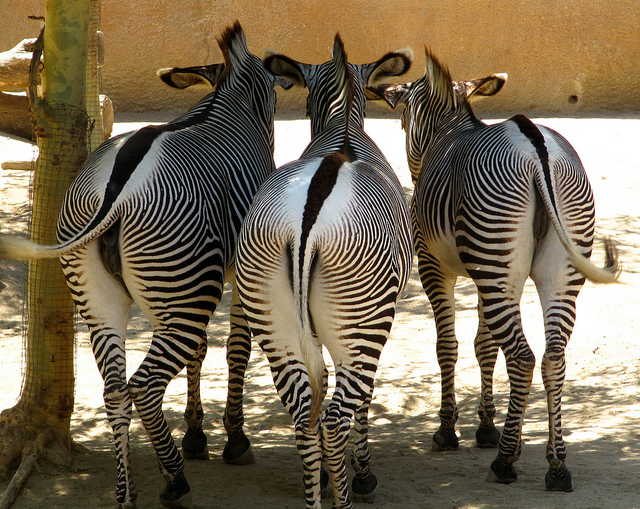} & zebra & Zebras standing together in a field with zebras & A group of zebras standing in a line & Zebra,Enclosure,Zoo \\
\hline
\end{tabular}
\end{table*}
\section{Conclusion and Future Work}
In this paper, we presented an approach to generate captions for images that lack parallel captions during training. Experimental results on unseen/novel image objects captioning exhibit that usage of structured information encapsulated 
in the form of relational knowledge (i.e KB) has unveiled a way to build connection between real world objects to their visual information. In future, we plan to expand our models to build 
multimedia knowledge bases that automatically can be queried based on relational information between images.

\appendix
\section{Appendix}
This appendix provides information on some of the quantitative and qualitative results of individual components in KGA-CGM. Also, more qualitative results of the generated 
captions on held-out MSCOCO objects is presented. 

\subsection{Language Model Hidden Layers}
As presented in the paper, language model in our caption generation model (i.e KGA-CGM) is a 2-layer forward LSTM. For learning KGA-CGM with image-caption pairs, input caption word embeddings 
are chosen to be 256 dimensions, while the LSTM hidden layer dimensions for both layer-1 and layer-2 is selected as 512. However, varying hidden layer dimensions can show an influence on the 
caption generation results. In this section, we vary the hidden layer dimensions and analyze the consequences. Table~\ref{capgen-hidden} shows the METEOR, SPICE and F1 average measures on 8 
unseen MSCOCO objects.
\begin{table}[!ht]
 \centering
 \small
 \caption{Effect on KGA-CGM with varying LSTM hidden layer dimensions in Language model.}
 \label{capgen-hidden}
 \begin{tabular}{ l|l|l|l|l|l}
 \hline
 Layer-1 & Layer-2 & Beam &  METEOR & SPICE & F1-score \\
 \hline
  256 & 256 & 1 & 20.9 & 13.5 &  50.8\\
  256 & 512 & 1  & 21.1 & 13.6 & 48.2 \\
  512 & 512 & 1  & \textbf {22.2} & \textbf{14.6} & \textbf{54.5} \\
 \cline{3-3}
  256 & 256 & $>$1 & 20.2 & 13.2 & 42.9 \\
  256 & 512 & $>$1  & 20.2 & 13.1 & 41.8 \\
  512 & 512 & $>$1  & 21.5 & 13.9 & 48.9 \\
  \hline
 \end{tabular}
\end{table}
\subsection{KGA-CGM More Qualitative Results}
\begin{figure*}
    \centering
    \begin{subfigure}[b]{0.35\textwidth}
        \includegraphics[width=\textwidth]{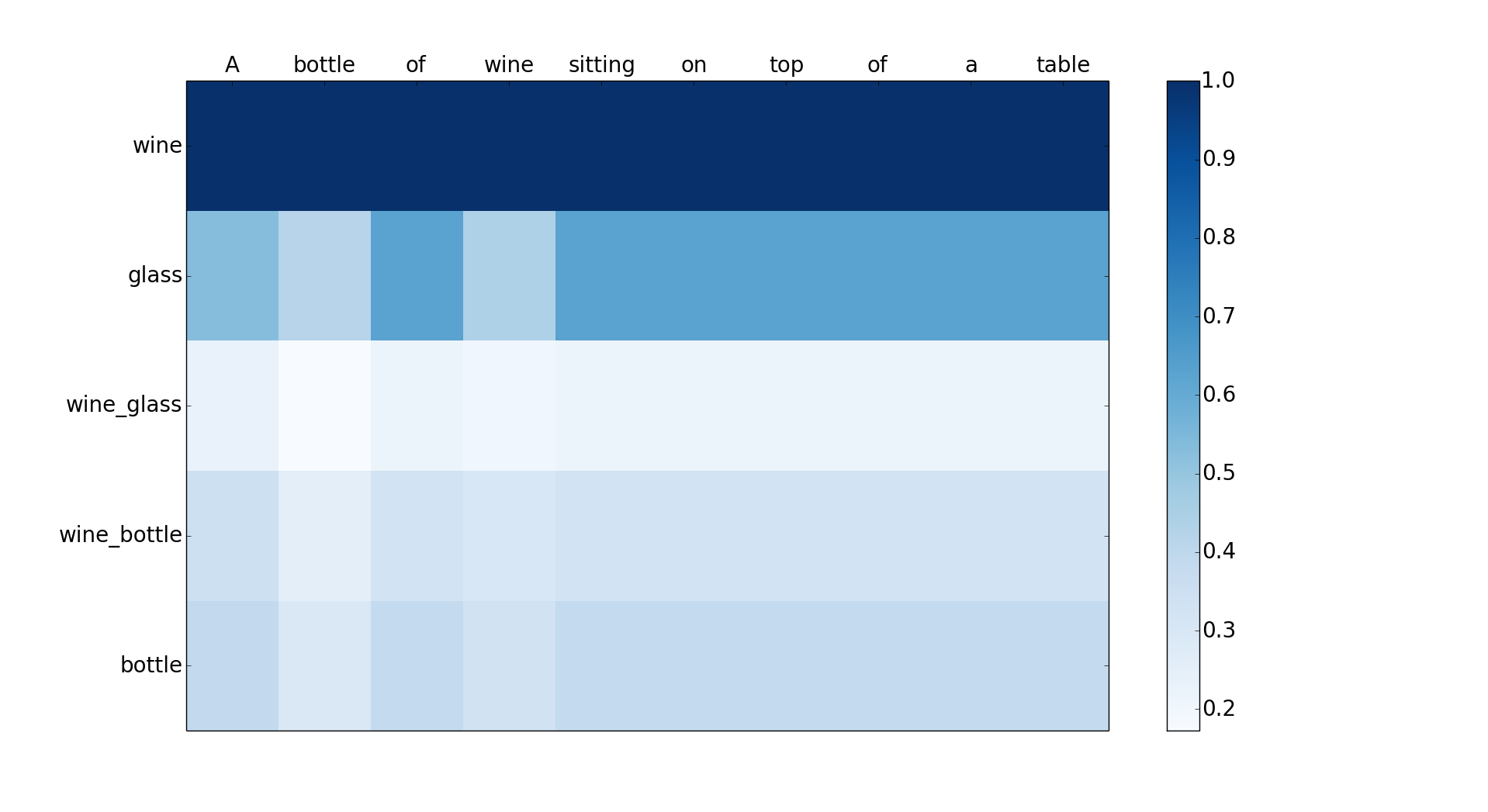}
        \caption{Bottle}
        \label{fig:bottle}
    \end{subfigure}
    ~ 
    \begin{subfigure}[b]{0.35\textwidth}
        \includegraphics[width=\textwidth]{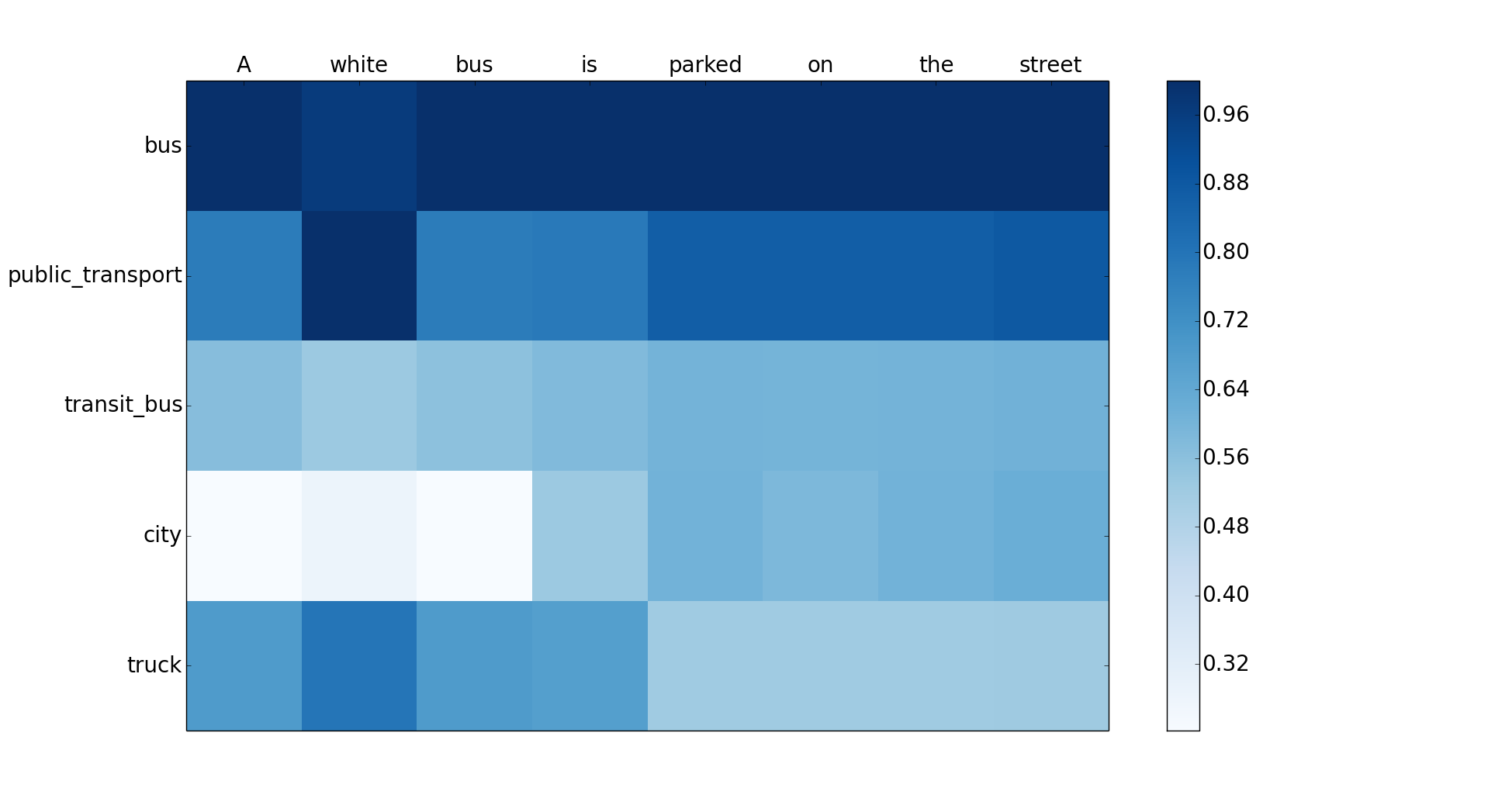}
        \caption{Bus}
        \label{fig:bus}
    \end{subfigure}
    ~ 
    \begin{subfigure}[b]{0.32\textwidth}
        \includegraphics[width=\textwidth]{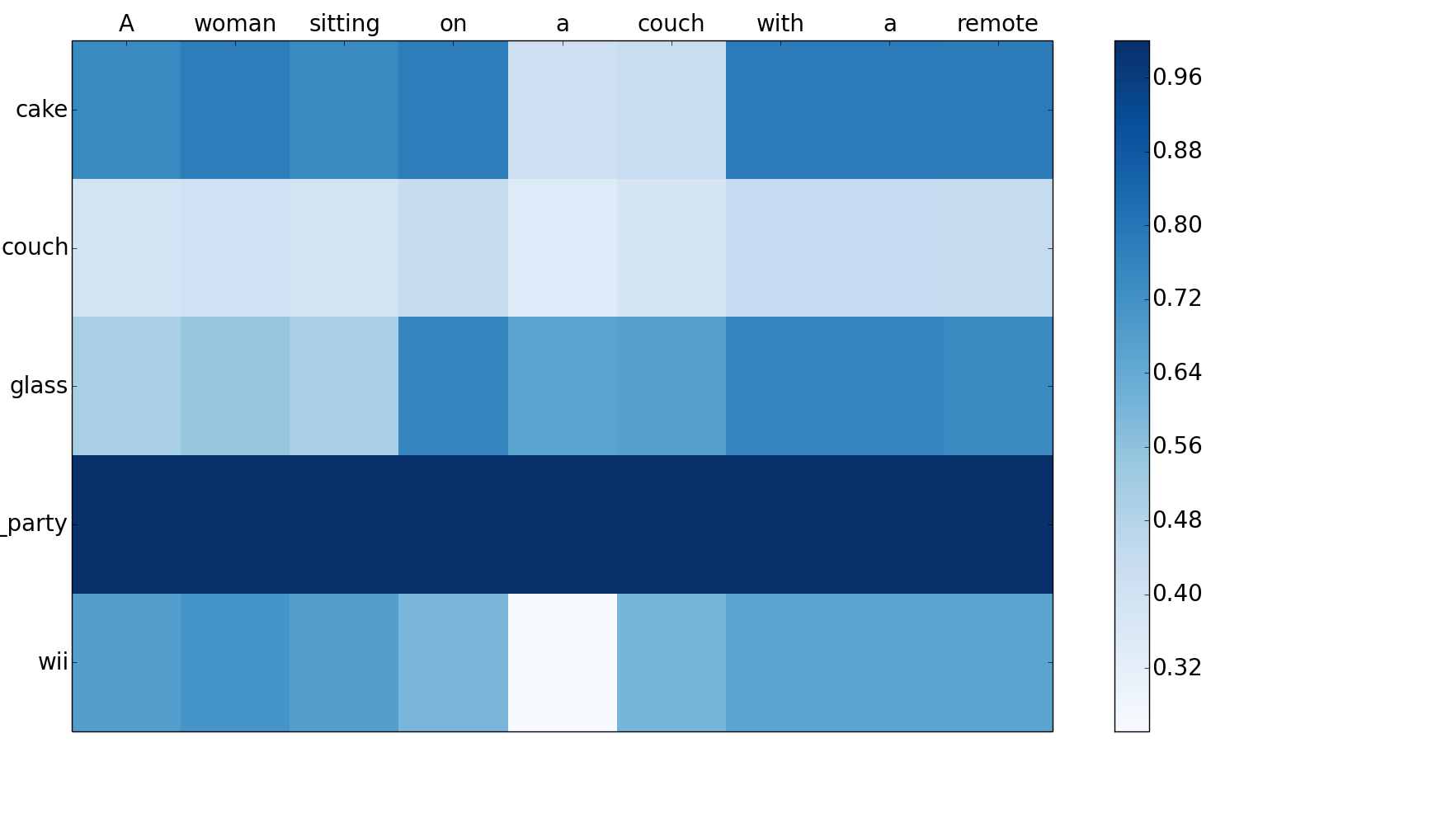}
        \caption{Couch}
        \label{fig:couch}
    \end{subfigure}
    \begin{subfigure}[b]{0.35\textwidth}
        \includegraphics[width=\textwidth]{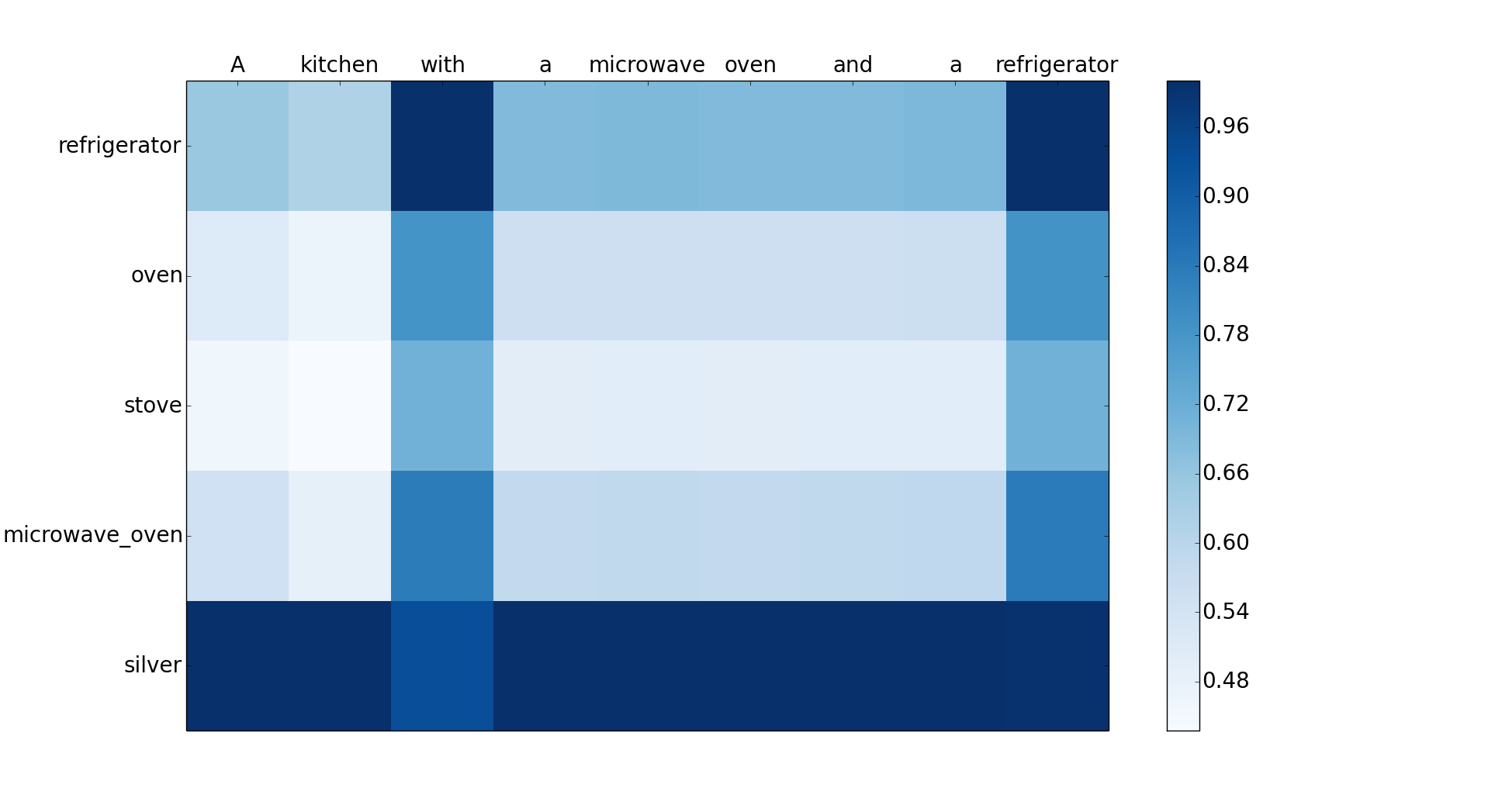}
        \caption{Microwave}
        \label{fig:microwave}
    \end{subfigure}
    \begin{subfigure}[b]{0.35\textwidth}
        \includegraphics[width=\textwidth]{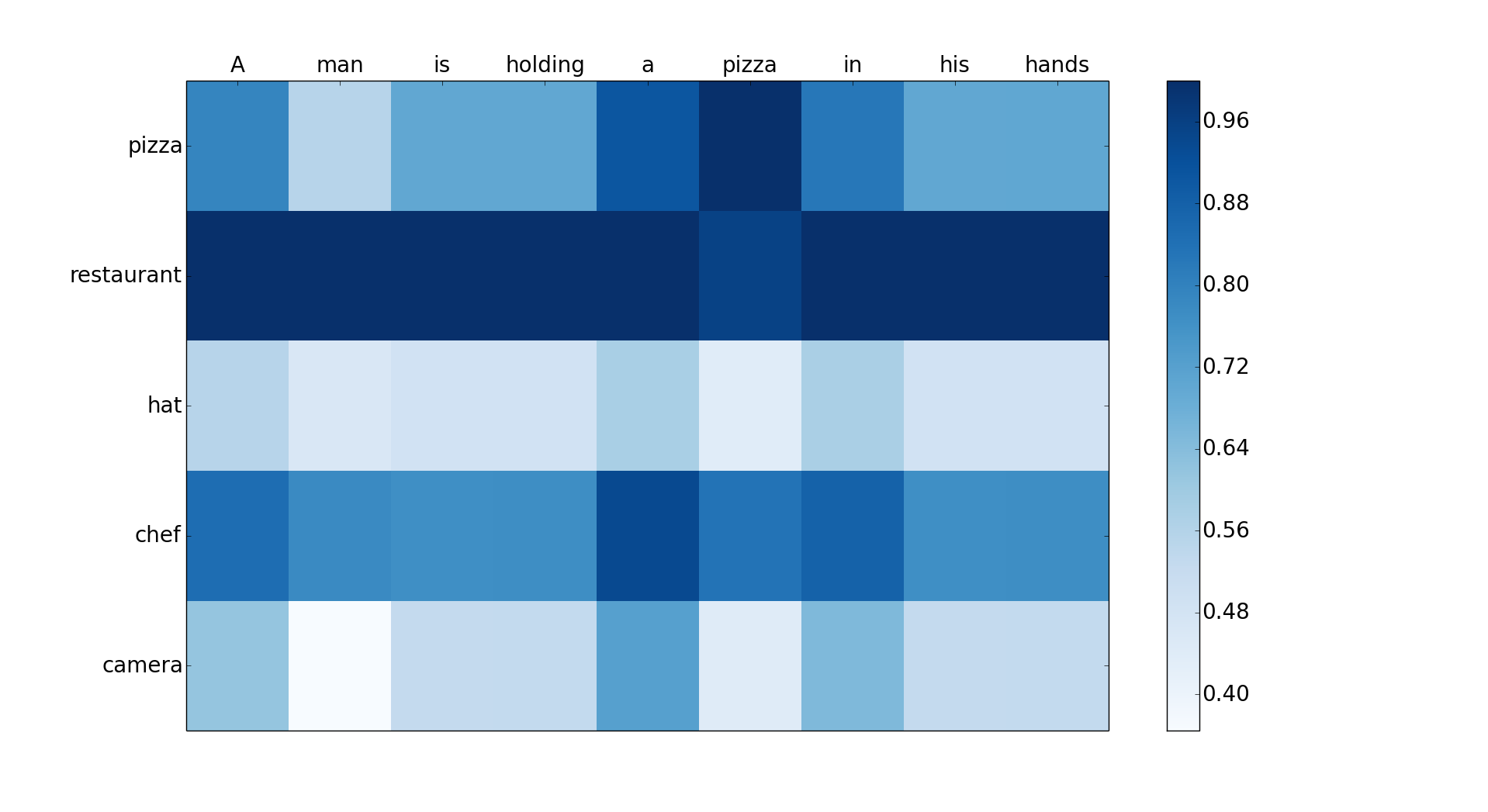}
        \caption{Pizza}
        \label{fig:pizza}
    \end{subfigure}
    ~ 
    \begin{subfigure}[b]{0.35\textwidth}
        \includegraphics[width=\textwidth]{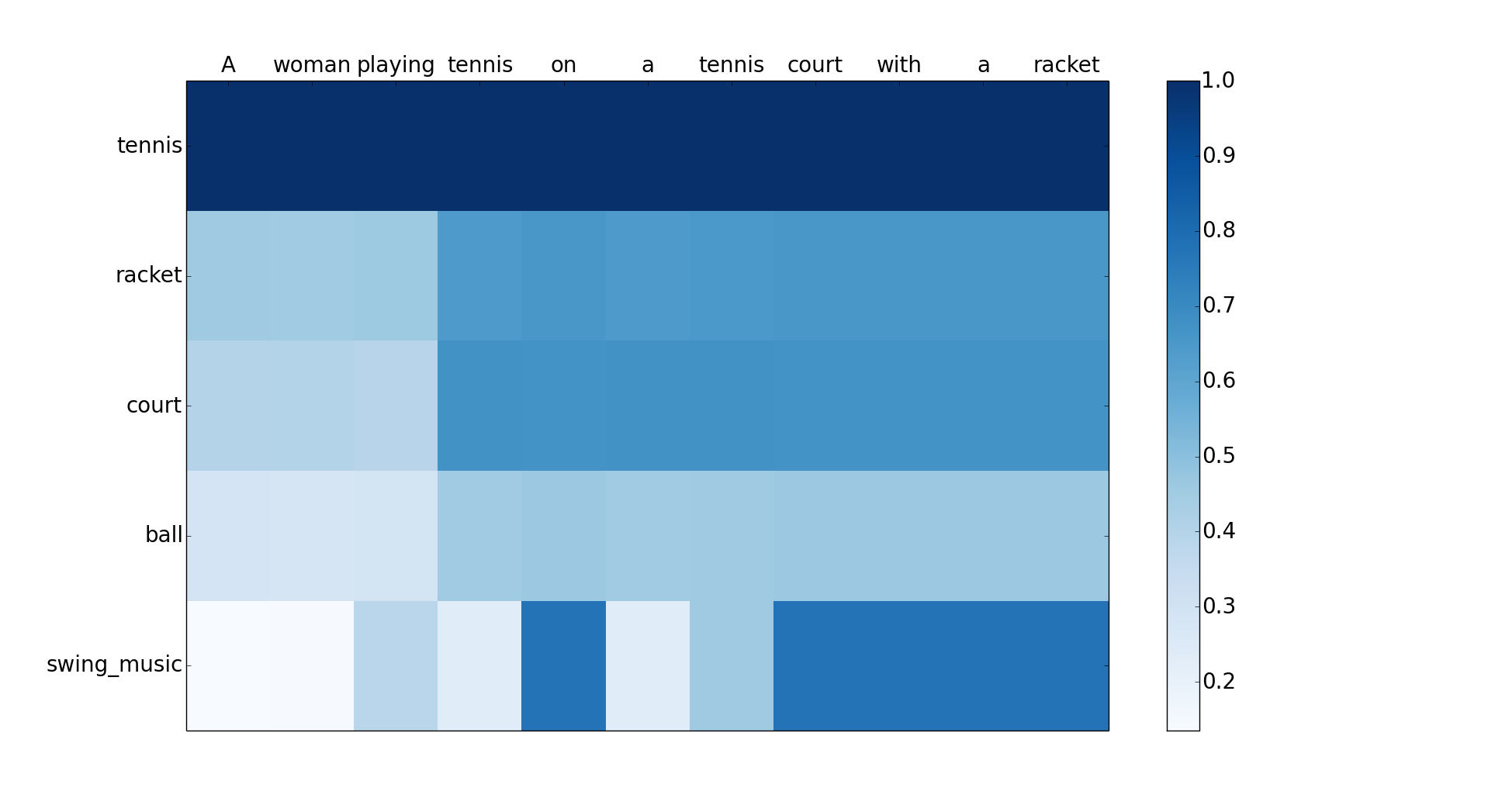}
        \caption{Racket}
        \label{fig:racket}
    \end{subfigure}
    ~ 
    \begin{subfigure}[b]{0.35\textwidth}
        \includegraphics[width=\textwidth]{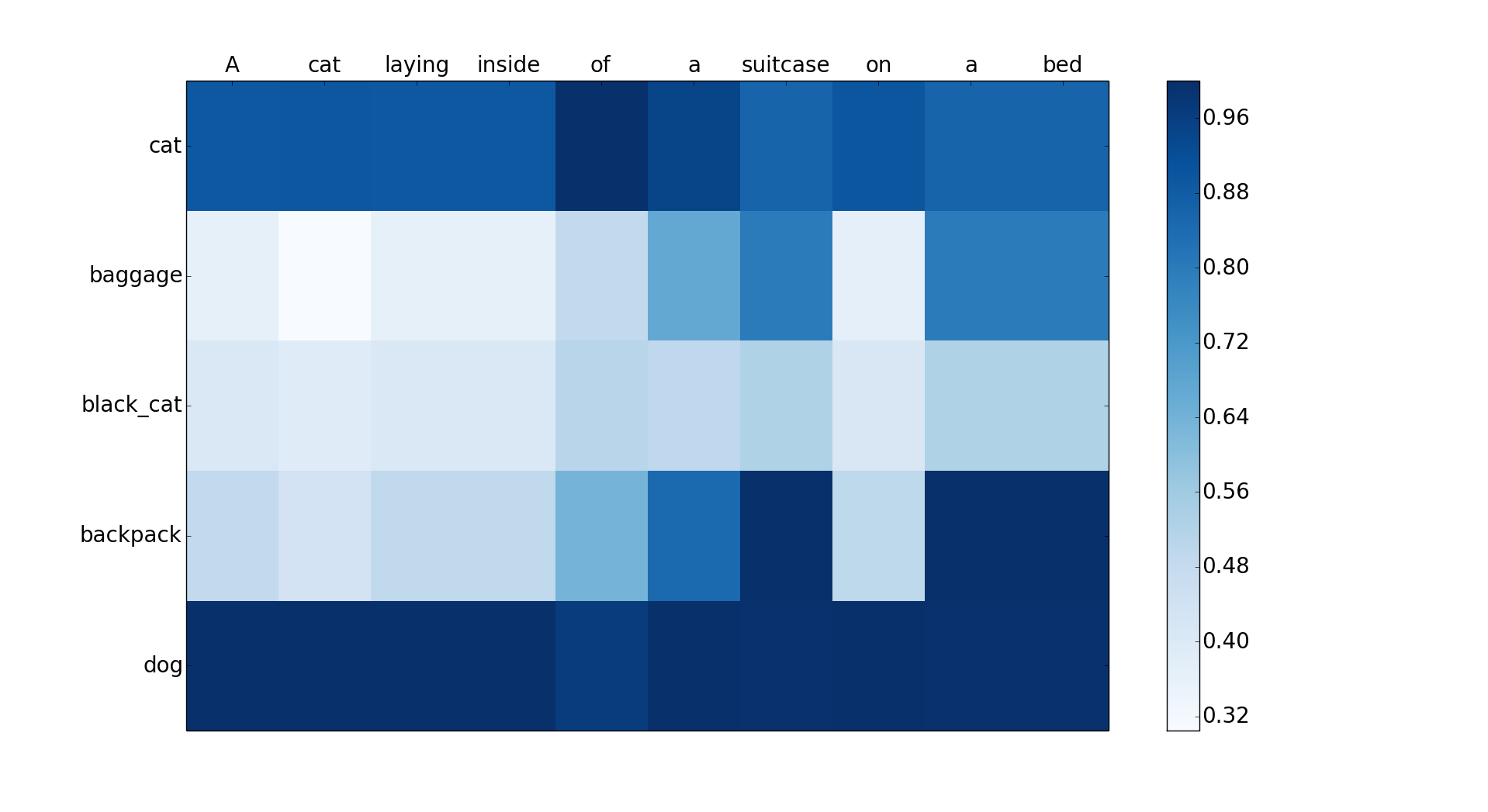}
        \caption{Suitcase}
        \label{fig:suitcase}
    \end{subfigure}
    \begin{subfigure}[b]{0.35\textwidth}
        \includegraphics[width=\textwidth]{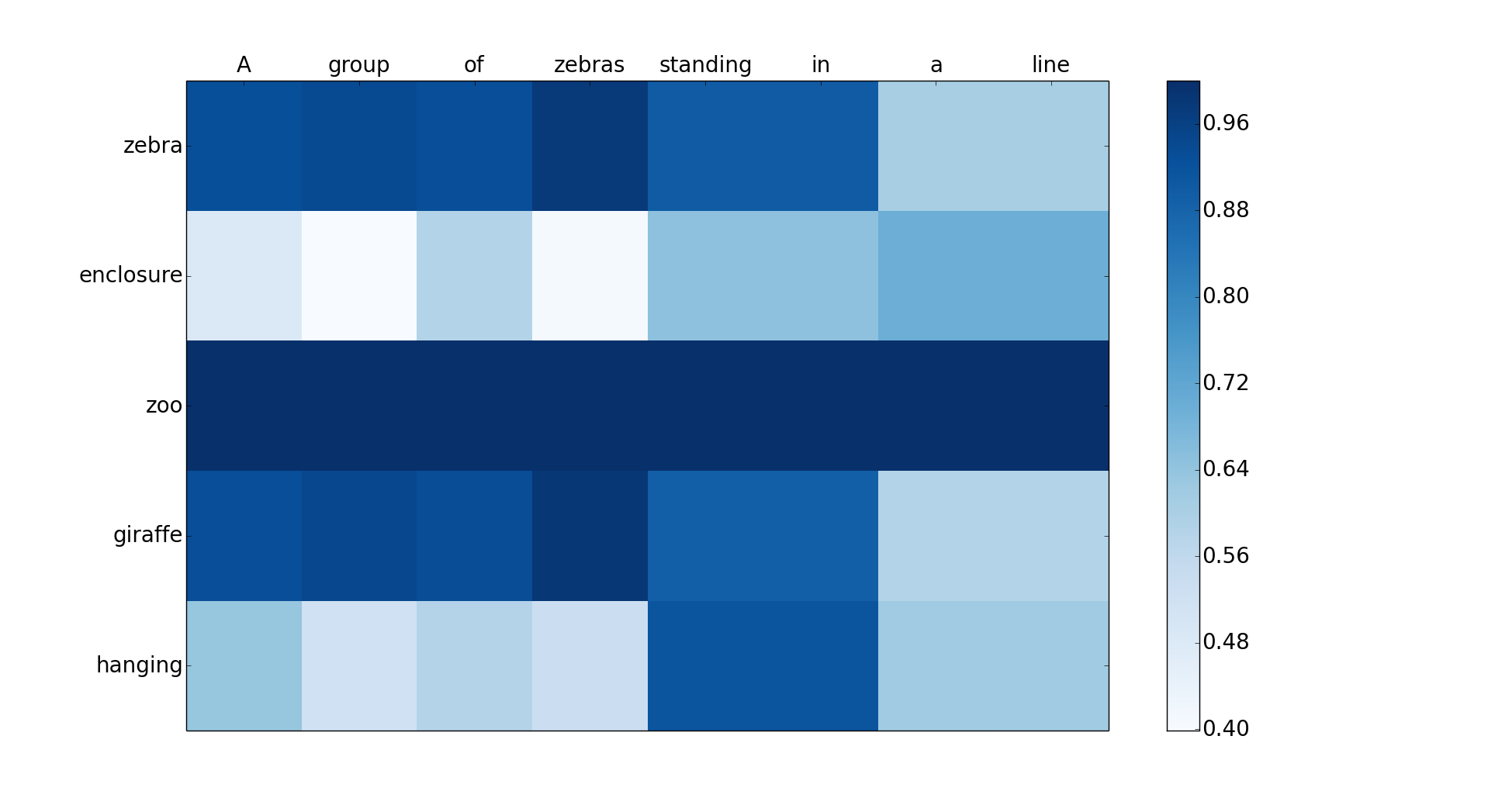}
        \caption{Zebra}
        \label{fig:zebra}
    \end{subfigure}
    \caption{Visualization of ESA attention weights of the captions presented in Table~\ref{kgareprpred}. X-axis labels represent generated captions and Y-axis labels are predicted 
    entity-labels. Scores are normalized between 0 and 1.}\label{fig:unseen}
\end{figure*}

The attention weights ($\boldsymbol{\beta}_{ti}$) of ESA contributes to the caption generation. Figure~\ref{fig:unseen} visualizes the attention weights of the  captions presented in the 
Table~\ref{kgareprpred}.

\bibliographystyle{ACM-Reference-Format}
\bibliography{sample-bibliography} 

\end{document}